\pdfoutput=1

\def\year{2020}\relax
\documentclass[letterpaper]{article} 
\usepackage{aaai20}  
\usepackage{times}  
\usepackage{helvet} 
\usepackage{courier}  
\usepackage[hyphens]{url}  
\usepackage{graphicx} 
\def\UrlFont{\rm}  
\usepackage{graphicx}  
\usepackage{comment}
\usepackage{natbib}
\usepackage{color}

\newcommand{\andre}[1]{\textcolor[rgb]{1.0, 0.4, 0.0}{[A: {\it #1}]}}

\title{SENSAR: A Visual Tool for Intelligent Robots for Collaborative\\ Human-Robot Interaction}

\author{Andre Cleaver\textsuperscript{\rm 1}, Faizan Muhammad\textsuperscript{\rm 2}, Amel Hassan\textsuperscript{\rm 2}, Elaine Short\textsuperscript{\rm 2}, Jivko Sinapov\textsuperscript{\rm 2} \\ 
\textsuperscript{\rm 1}Department of Mechanical Engineering\\ 
\textsuperscript{\rm 2}Department of Computer Science}

\begin{document}

\maketitle

\begin{abstract}
Establishing common ground between an intelligent robot and a human requires communication of the robot's intention, behavior, and knowledge to the human to build trust and assure safety in a shared environment. This paper introduces SENSAR (Seeing Everything iN Situ with Augmented Reality), an augmented reality robotic system that enables robots to communicate 
their sensory and cognitive data in context over the real-world with rendered graphics, allowing a user to understand, correct, and validate the robot's perception of the world. Our system aims to support human-robot interaction research by establishing common ground where the perceptions of the human and the robot align. 
\end{abstract}

\section{INTRODUCTION}
For intelligent robots to operate in the same environment as humans, they must communicate their intentions, behaviors, and knowledge in a way that is interpretable by humans. Robots rely on sensors and algorithms to support functions such as localization, obstacle avoidance, and object manipulation; however, robots can misidentify objects and communicate their misperceptions of the world, which can lead to undesired outcomes. As a result, humans may not understand what a robot is trying to do, with adverse effects on human-robot collaboration. One way to solve this problem is to allow the human collaborator to directly see a robot's perception to ensure the robot is performing adequately \citep{thorstensen_visualization_2017}. 

Augmented Reality (AR) renders spatially-grounded images over the real world. By leveraging AR, robot data that would typically be ``hidden'' can be visualized by rendering graphic images over the real world using AR-supported devices (e.g., smartphones, tablets, or the Microsoft HoloLens) \citep{frank2017mobile}.
For example, a robot's depth sensor data rendered directly over a workstation shows not only what the robot ``sees'' but also what the robot does \emph{not} ``see'': that is, it allows the human to see what the robot could not detect. AR gives humans the advantage of placing the robot's sensory data in the context of the real world unlike traditional visual programs (e.g., rviz\footnote{http://wiki.ros.org/rviz}).

\begin{figure}[t!]
\centering
\includegraphics[width=0.78\linewidth]{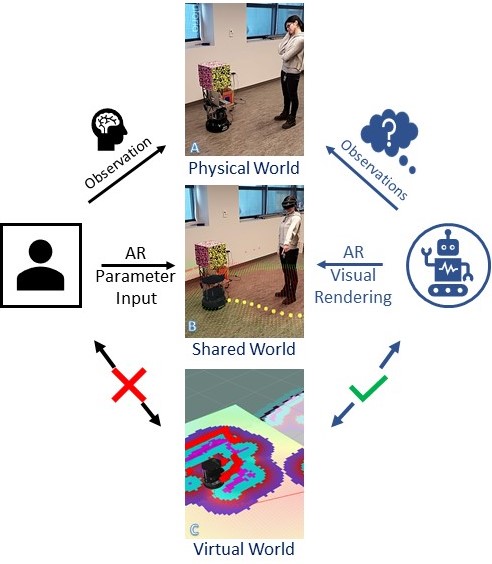}
\caption{(Right) Physical world known by a human; robots are unaware of most objects within environment, (Left) Virtual World accessed only by the robot what includes the robot's algorithms and sensor readings, (Center) a shared reality world accessed by both humans and robots. A robot renders its path plan using markers which is seen by human.}
\label{fig:sensor_figure}
\end{figure}

In this paper, we introduce SENSAR (Seeing Everything iN Sitiu with Augmented Reality), a novel AR-mediated communication tool. We describe the design of our system, which uses ROS and Unity, along with the data processing routines to convert raw data to visuals. Our system can create an interactive reality space containing the robot's sensory data and cognitive output that a user can access to reach a ``common ground'' with the robot (see Figure \ref{fig:sensor_figure}). In a technical demonstration of the system, we show that visualizing sensors and intended actions can be beneficial for humans in areas such as debugging, education, and serve as a tool for HRI studies.


\section{RELATED WORK}

Currently, there are frameworks that already exist that display robot data using external lights and projections \citep{fernandez2018passive,chadalavada2015s}. Although these methods have the advantage of not requiring the user to hold or wear a separate device;

they are not without limitations. These solutions require modifying the robot with hardware and staging the environment with flat surfaces to view the projections. Occlusions are another inevitable problem with this approach if any objects or humans themselves are positioned between the display surface and the projection source. AR does not face this problem as graphical images are rendered over the real-world, thus eliminating the need for any changes to the existing robot or environment. 

AR has been integrated with several robotic platforms with a focus in areas such as navigation, programming, and education \citep{walker2018communicating,gadre2019end,cheli2018towards}. Work by \citet{cheli2018towards} explored AR as an education tool for K-12 students. Middle school students were observed debugging their assigned robots (EV3 Kit) through Tablets and initiated group discussions around sensor readings. The EV3 kit contained actuators and sensors and such as touch, color detection, sonar, and gyro. Although their system was successful in visualizing the robot's data, the system can only function with the EV3 robot. We demonstrate SENSAR with a Turtlebot2 robot, a popular robotic platform for research; however, any ROS-based robot can work with the SENSAR system. In addition, SENSAR is highly modular: multiple sensors including the type of sensors provided in the EV3 kit can be connected to the robot such as depth and Lidar sensors. SENSAR is a continuation of prior work by \citet{muhammad_creating_2019}.

\section{SYSTEM ARCHITECTURE AND DESIGN}
SENSAR is an AR system\footnote{The SENSAR package repositories can be found at: \url{https://github.com/tufts-ai-robotics-group/arfuros_ros.git} and \url{https://github.com/tufts-ai-robotics-group/arfuros.git}} enabling users to ``see into the mind'' of a robot. Data collected by an intelligent robot is processed through various filtering techniques (e.g., downsampling) to render the output in a visual form that is intuitive for a user. The functional requirements of SENSAR included:
\begin{itemize}
    \item Customizable for different sensors and common data structures within a robotics system
    \item Operational without external tracking or cloud service
    \item Versatile to multiple AR-supported devices
  
\end{itemize} 


\subsection{Hardware and Software}
SENSAR requires two components: an intelligent robot and an AR device.
A \textit{Turtlebot2} served as the intelligent robot; it is controlled through the Robot Operating System (ROS) \citep{quigley2009ros} running on Ubuntu 16.04. The robot is equipped with an \textit{RPLIDAR A2M8 Laser Scanner} and an \textit{Astra Camera} depth sensor. The robotic sensors and cognitive data are extracted and filtered from ROS using Python and C++ scripts. For the AR device, we used a Microsoft Hololens, a Samsung Galaxy S9 Android smartphone device with a 12 MP Camera, and an iPad to render visuals over the real-world. Our system can be installed on any smartphone, iPhone, or tablet device that meets the hardware requirements to support AR applications. 

The AR device utilizes \textit{Unity}\footnote{\url{https://unity.com/}} to create visualizations to project data around the robot. \textit{Vuforia}\footnote{\url{https://developer.vuforia.com/}} is used to establish physical tracking of the robot using a target image. Once the pose of the robot within the incoming video feed is detected, the data received from the robot can then be projected onto the camera feed. In ROS, \textit{ROSBridge} provides a JSON interface to the topics native to ROS, whereas in Unity, \textit{ROS Sharp} provides C\# functions and classes to interact with \textit{ROSBridge} through JSON. Through a WebSocket connection, ROS messages of any type from the robot come out as native C\# data structures at the other end of the pipeline.

\begin{figure}[t!]
\centering
\includegraphics[width=.90\linewidth]{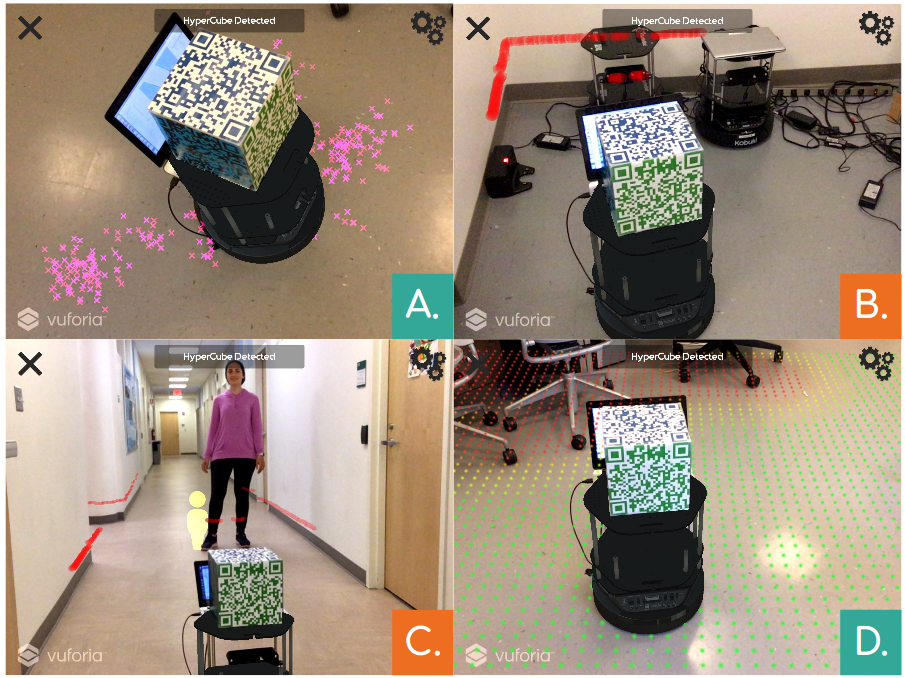}
\caption{This figure shows the first design of the robot and the types of visual data you display with AR. (A) Localization, (B) Laser scan, (C) People Detection, (D) Cost Map.}
\label{fig:visuals}
\end{figure}

\subsection{Visualization Options}
Below is a list of current visualizations that are equipped with the SENSAR system. SENSAR is an ongoing project and we plan to add additional visualizations in the future.

\subsubsection{Localization Particles}
This data type is based on the Monte Carlo localization algorithm or \textit{Particle Filter Localization} that robots use to localize themselves within a given map \citep{MooreStouchKeneralizedEkf2014}. Figure \ref{fig:visuals}.A show markers in pink visualizing the robot's belief about its location within a given map. 

\subsubsection{Laser Scan}
A 2D LIDAR reading is shown in figure \ref{fig:visuals}.B Here, the laser scan detects the walls and objects with red markers. 

\subsubsection{Human Detection}
A human detection algorithm leverages the Laser Scan data as input and compares the measurements against the profile of a pair of human legs. Legs are used as a target feature due to the placement of the LIDAR sensor. A human avatar is rendered in place of a coordinate data point that represents a detected human. Figure \ref{fig:visuals}.C shows the Human Detection visual as seem by the yellow human avatar. 

\subsubsection{Occupancy Grid (Cost Map)}
Cost Map is a measure of occupancy within the surrounding area of a robot. The robot reports a planar grid consisting of cells using sensory data and information from a static map. Each cell represents a probability of occupancy [0-1]. The information for each cell influences the robot's path trajectory by selecting cells with low probability of containing an obstacle. For visualization, we related the probability value of a cell to correspond to a color. Figure \ref{fig:visuals}.D shows the \textit{Cost Map}. Here, the robot detects the chairs in its surroundings and then renders red markers (high probability of occupancy) in their location and green markers for areas that are clear.

\subsubsection{Path Trajectory}
Path Trajectory visual represents the robot's future trajectory to a selected destination point as a series of markers on the ground plane. The trajectory is generated by the robot's internal path planner algorithm. Figure \ref{fig:navigation_visuals} Top shows the robot's  path trajectory around the obstacle (green plushie).

\subsubsection{Turning Signal}
\begin{figure}
\centering
\includegraphics[width=0.75\linewidth]{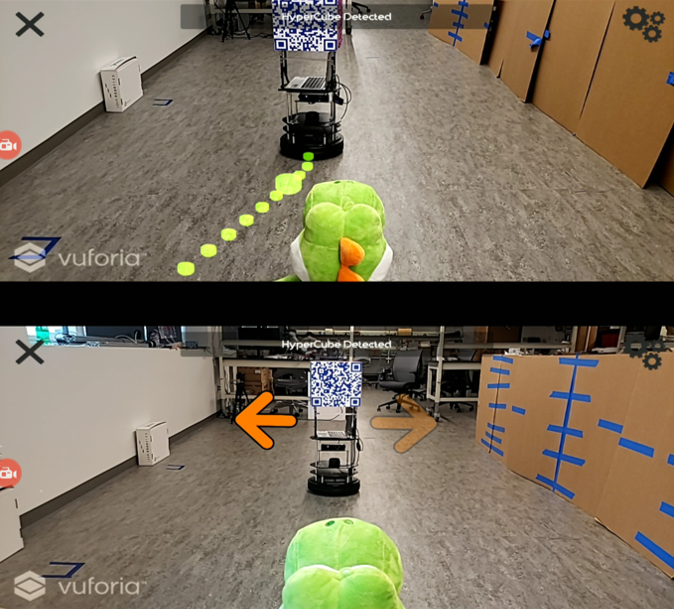}
\caption{Visualization projected by the robot to indicate intended direction of motion around an obstacle. Top image is path trajectory using yellow markers. Bottom image is turn signals which operates similar to turn signals of a motor vehicle.}
\label{fig:navigation_visuals}
\end{figure}

The Turning Signal data  type alerts a user on the robot's intended direction of navigation by flashing an arrow towards the respective direction. Turn Signal mimics a motor vehicle's turning signal system. Figure \ref{fig:navigation_visuals} Bottom shows the robot's flashing an arrow towards the direction it will move.

\subsubsection{Depth Camera Sensor (\textit{PointCloud)}}
PointCloud renders the depth sensor data as particles in 3D space. Filtering and segmentation techniques are applied for model detection algorithms.

\section{APPLICATIONS}

\subsection{Conveying Navigational Intent}
We are exploring how SENSAR can leverage the Path Trajectory and Turning Signal data types for users to interpret the robot’s motion intent. These navigation visuals will be evaluated using videos clips. Participants will watch segments of the robot approaching an obstacle. The robot will travel either to the left or right around an obstacle as depicted in figure \ref{fig:navigation_visuals}, and participants will select the direction they believe the robot will travel towards. 

\subsection{Robotic Education for Kids}
\begin{figure}
\centering
\includegraphics[width=0.4\textwidth]{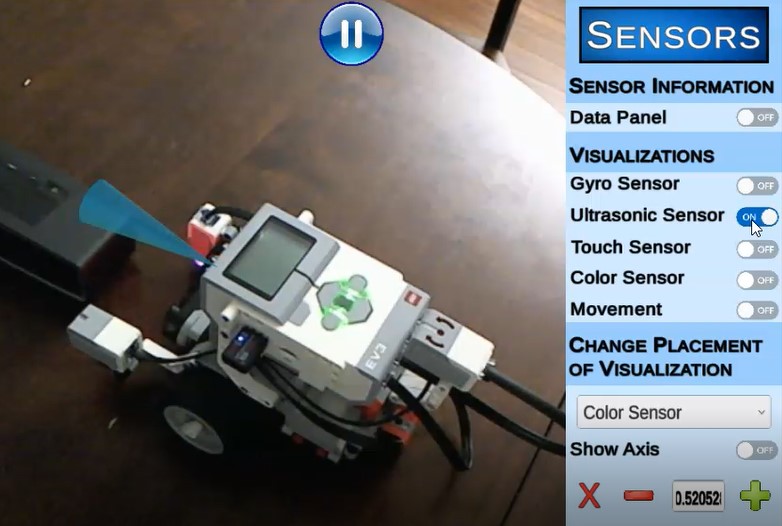}  
\caption{EV3 robot is projecting its ultrasonic sensor data by visualizing the blue cone that extends out towards the object detected by the robot's sensor. Other sensor can be visualized as seen on the right-hand panel.}
\label{fig:ev3throughts}
\end{figure}

we are exploring how an AR robotic platform performs as a teaching tool for robotic education for kids. We applied our SENSAR framework to work with an EV3 robot as seen in figure \ref{fig:ev3throughts}. Here, the EV3 robot contains additional sensors that can be projected to give kids a visual aid on key concepts to robots while interacting with their robot. The next steps are to deploy our system and observe how students in small groups interact with the novel system and collaborate with each other to accomplish learning objectives.

\begin{figure*}[t!]
\centering
\includegraphics[width=0.95\textwidth]{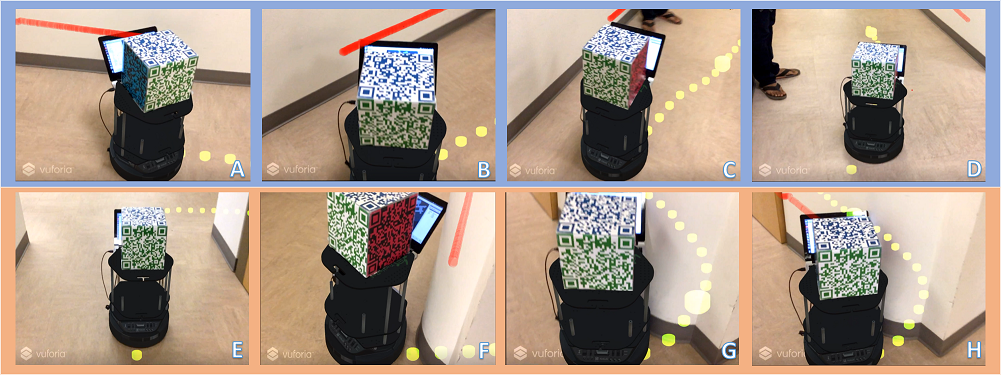}  
\caption{Planned path trajectory (yellow markers) with operating laser scan (Red markers) resulting in obstacle avoidance (top row). Robot operating with unexpected malfunctioning laser scan leading to wall collision (bottom row). laser scan did not detect the wall as the robot planned a right turn in image E. The lack of red markers in image E but present in the following image F provided insight that the laser scan is the source of the robot's behavior.}
\label{fig:path_malfunction}
\end{figure*}

\subsection{Robotic Debugging}
Figure \ref{fig:path_malfunction} shows a sequence of images during two cases when the robot was navigating through a building to reach a specific destination. In the top set of images (A-D), the robot operated with a functioning LIDAR sensor, as seen by the red markers detecting the nearby wall. The path-planner algorithm incorporates the laser scan data with the map of the building to determine available routes to the destination while avoiding obstacles. Here, the robot successfully avoids the wall to continue down the hallway. In the bottom set of images (E-H), the robot operated with a malfunctioning LIDAR sensor, as evidenced by the absence of red markers in image E indicating that the wall is not detected and therefore is not an obstacle for the robot. As a result, the robot re-calculates a new shorter path, and the robot thus collides into the wall. SENSAR provided the user with insight into the reason for the collision, which would have been difficult to determine if the user had only visually observed the robot.


\section{CONCLUSIONS}
This paper presented SENSAR, a novel AR-based robotic tool that enables robots to visually communicate their cognitive and sensory data. The major significance of this project is introducing a new and attractive form of interaction in HRI. By translating complex data into intuitive visuals, we can enable humans of all levels of expertise to work with robots. We described SENSAR's open-source hardware and software components which are available online through GitHub repositories. We then describe domains that our system may benefit. We expect our system to become a useful tool for other HRI researchers investigating how and whether visualizing various robot data in AR can improve human-robot collaboration and enable effective learning and reasoning.



\bibliographystyle{aaai}
\bibliography{references}

\end{document}